\documentclass{article}

\usepackage{ijcai20}

\usepackage{amsmath,amssymb}
\usepackage{color}
\usepackage{tabu} 
\usepackage{stackengine}

\usepackage{dsfont}

\usepackage{natbib}

\usepackage{caption}
\usepackage{subcaption}
\usepackage{dblfloatfix}

\definecolor{darkgray}{rgb}{0.66, 0.66, 0.66}

\usepackage{authblk}

\usepackage{verbatim}
\usepackage{algorithm}
\usepackage[noend]{algpseudocode}
\usepackage{microtype}
\usepackage{graphicx}
\usepackage{booktabs} 
\usepackage{wrapfig}

\usepackage[colorinlistoftodos]{todonotes}

\usepackage{hyperref}

\DeclareMathOperator*{\argmax}{arg\,max}

\DeclareMathOperator*{\argmin}{arg\,min}

\DeclareMathOperator{\bigO}{\mathcal{O}}
\DeclareMathOperator{\reals}{\mathbb{R}}
\DeclareMathOperator{\integers}{\mathbb{N}}
\DeclareMathOperator{\state}{\mathbf{s}}
\newtheorem{theorem}{Theorem}
\newtheorem{definition}{Definition}

\newcommand{\xx}{\mathbf{x}}
\newcommand{\aaa}{\mathbf{a}_\text{env}}

\newcommand{\dd}{\mathbf{d}}
\newcommand{\yy}{\mathbf{y}}

\newcommand{\DD}{\mathbf{D}}
\newcommand{\XX}{\mathbf{X}}

\newcommand{\aae}{\mathbf{a}_\text{SoE}}
\newcommand{\aab}{\mathbf{a}_\text{BoW}}

\usepackage{xargs}
\newcommandx{\done}[2][1=]{\todo[inline,linecolor=blue,backgroundcolor=blue!25,bordercolor=blue,#1]{#2}}

\usepackage{pifont} 
\newcommand{\cmark}{\ding{51}}%
\newcommand{\xmark}{\ding{55}}%
\usepackage{array}\usepackage{makecell} 

\newcommand{\algfull}{Sparse Imitation Learning}
\newcommand{\alg}{Sparse-IL}
\usepackage[capitalize]{cleveref}

\usepackage{capt-of}

\begin{document}

\title{Action Assembly: Sparse Imitation Learning for Text Based Games with Combinatorial Action Spaces}
\author[*,1]{Chen Tessler}
\author[*,3]{Tom Zahavy}
\author[2]{Deborah Cohen}
\author[3]{Daniel J. Mankowitz}
\author[1]{Shie Mannor}
\affil[*]{Equal contribution}
\affil[1]{Technion Israel Institute of Technology, Haifa, Israel}
\affil[2]{Google Research, Tel-Aviv, Israel}
\affil[3]{DeepMind, Longdon, England}
\date{}                     
\setcounter{Maxaffil}{0}
\renewcommand\Affilfont{\itshape\small}

\maketitle

\begin{abstract}
We propose a computationally efficient algorithm that combines compressed sensing with imitation learning to solve text-based games with combinatorial action spaces. Specifically, we introduce a new compressed sensing algorithm, named IK-OMP, which can be seen as an extension to the Orthogonal Matching Pursuit (OMP). We incorporate IK-OMP into a supervised imitation learning setting and show that the combined approach (Sparse Imitation Learning, Sparse-IL) solves the entire text-based game of Zork1 with an action space of approximately 10 million actions given both perfect and noisy demonstrations.

\end{abstract}


\section{Introduction}

Combinatorial action spaces pose a challenging problem for AI agents, both from a computational and from an exploratory point of view. The reason being that (i) finding the best action may require iterating over all actions, an exponentially hard task, and (ii) absent prior knowledge, finding the best action requires testing all actions multiple times at each state \citep{brafman2002r}. While the exploratory task is of great importance, in this work we focus on the computational aspects of the problem. Our method can be seen as a natural application of the Action Assembly Theory (AAT) \citep{greene2008action}. According to Greene, behavior is described by two essential processes: \textit{representation} and \textit{processing}. Representation refers to the way information is coded and stored in the mind, whereas processing refers to the mental operations performed to retrieve this information \citep{greene2008action}. Having good representations of information and an efficient processing procedure allows us to quickly exploit highly rewarding nuances of an environment upon first discovery. 

\begin{figure}[t]
    \centering
    \includegraphics[width=\linewidth]{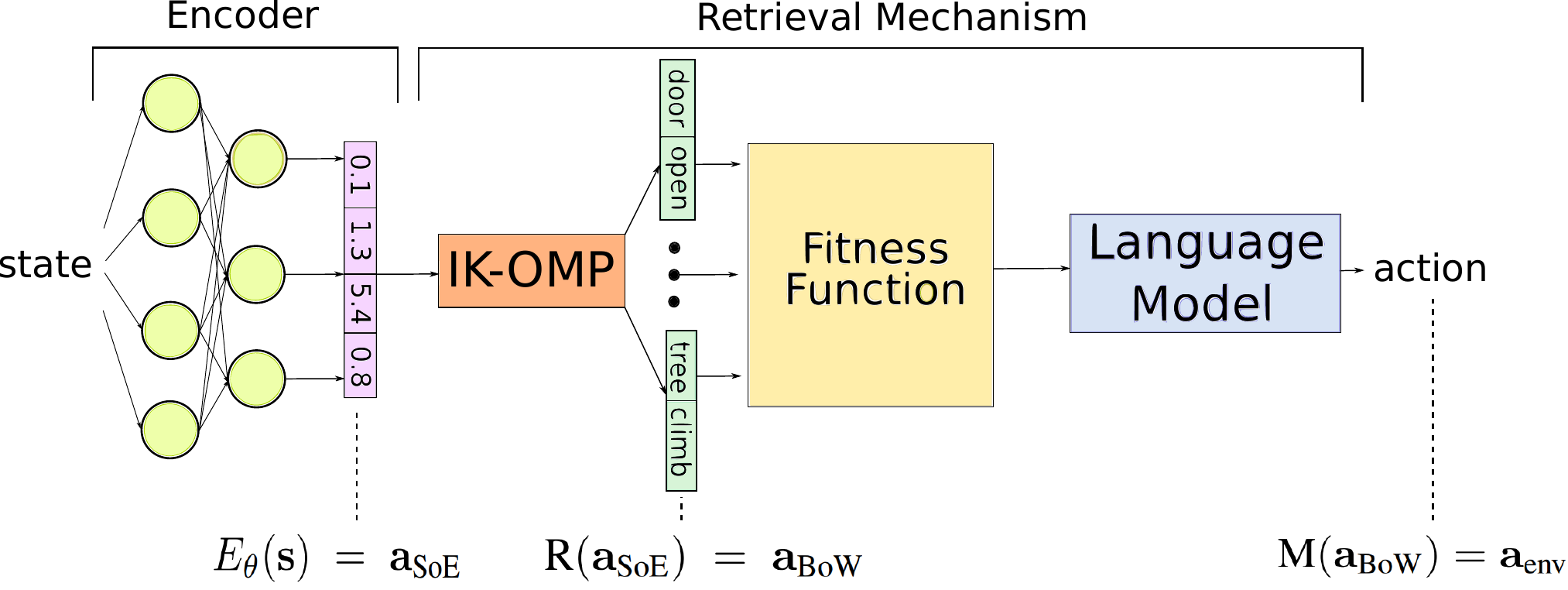}
    \caption{The \alg\ algorithm.}\label{fig:algorithm}
\end{figure}

In this work we propose the first computationally efficient algorithm (see Figure \ref{fig:algorithm}), called \textit{\algfull\ (\alg)}, which is inspired by AAT and combines imitation learning with a Compressed Sensing (CS) retrieval mechanism to solve text-based games with combinatorial action spaces. 
Our approach is composed of:


\textbf{(1) Encoder} - the encoder receives a state as input (Figure~\ref{fig:algorithm}). The state is composed of individual words represented by word embeddings that were previously trained on a large corpus of text. We train the encoder, using imitation learning, to generate a continuous action $\aae$ (a dense representation of the action). The action $\aae$ corresponds to a sum of word embeddings of the action that the agent intends to take, e.g., the embedding of the action `take egg' is the sum of the word embedding vectors of `take' and `egg'. As the embeddings capture a prior, i.e., similarity, over language, it enables improved generalization and robustness to noise when compared to an end-to-end approach.

\textbf{(2) Retrieval Mechanism} - given a continuous vector $\aae$, we reconstruct the $K$ best Bag-of-Words (BoW) actions $\aab$, composed of up to $l=4$ words, from the continuous output of the encoder. We do this using an algorithm that we term Integer K-Orthogonal Matching Pursuit (IK-OMP). We then use a fitness function to score the actions, after which, the best action is fed into a language model to yield an action sentence $\aaa$ that can be parsed by the game.

\textbf{Main contributions:} We propose a computationally efficient algorithm called Sparse-IL that combines CS with imitation learning to solve natural language tasks with combinatorial action spaces. We show that IK-OMP, which we adapted from \cite{white2016generating} and \cite{lin2013kbest}, can be used to recover a BoW vector from a sum of the individual word embeddings in a computationally efficient manner, even in the presence of significant noise. We demonstrate that \alg\ can solve the 
entire game of Zork1, for the first time, while considering a combinatorial action space of approximately 10 million actions, using noisy, imperfect demonstrations.

This paper is structured as follows: Section \ref{sec:relatedwork} details relevant related work. Section \ref{sec:problemsetting} provides an overview of the problem setting; that is, the text-based game of Zork and the challenges it poses. Section \ref{sec:compressed_sensing} provides an overview of CS algorithms and, in particular, our variant called IK-OMP. Section \ref{sec:imitationlearning} introduces our \alg\ algorithm. Finally, in Section \ref{sec: experiments} we present our empirical evaluations, which include experiments in the text-based game Zork1 highlighting the robustness of IK-OMP to noise and its computational efficiency and showcasing the ability of \alg\ solving both the `Troll Quest' and the entire game of Zork.

\section{Related work}
\label{sec:relatedwork}

\textit{Combinatorial action spaces in text-based games:} Previous works have suggested approaches for solving text-based games \citep{he2016deep,yuan2018counting,zahavy2018learn,zelinka2018using,tao2018towards}. However, these techniques do not scale to combinatorial action spaces. For example, \cite{he2016deep} presented the DRRN, which requires each action to be evaluated by the network. This results in a total of $\bigO(|A|)$ forward passes. \cite{zahavy2018learn} proposed the Action-Elimination DQN, resulting in a smaller action set $\bigO(|A'|)$. However, this set may still be of exponential size. 

\textit{CS and embeddings representation:} CS was originally introduced in the Machine Learning (ML) world by \cite{calderbank2009compressed}, who proposed the concept of compressed learning. That is, learning directly in the compressed domain, e.g. the embeddings domain in the Natural Language Processing (NLP) setting. The task of generating BoW from the sums of their word embeddings was first formulated by \cite{white2016generating}. A greedy approach, very similar to orthogonal matching pursuit (OMP), was proposed to iteratively find the words. However, this recovery task was only explicitly linked to the field of CS two years later in \cite{arora2018compressed}. 



\begin{figure}[h]
    \centering
    \includegraphics[width=\linewidth]{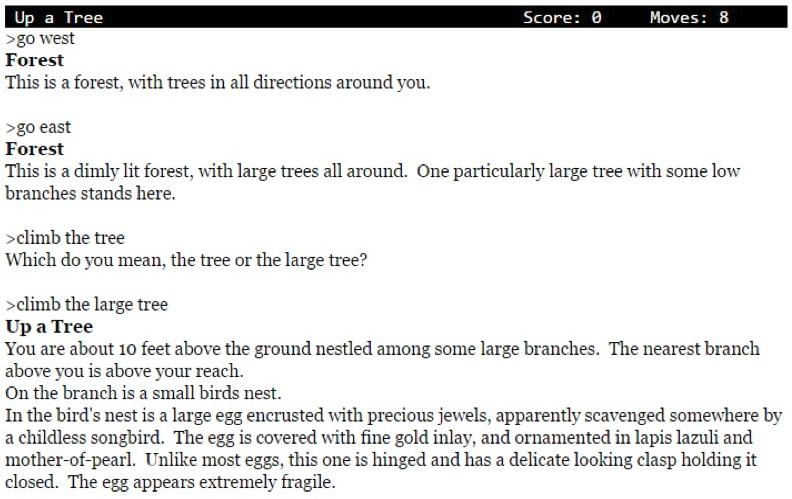}
    \caption{Zork1 example screen.}\label{fig:zork1_egg}
\end{figure}

\section{Problem setting}
\label{sec:problemsetting}

\paragraph{Zork - A text-based game:}
Text-based games \citep{cote2018textworld} are complex interactive games usually played through a command line terminal. An example of Zork1, a text-based game, is shown in Figure \ref{fig:zork1_egg}. In each turn, the player is presented with several lines of text which describe the state of the game, and the player acts by entering a text command. In order to cope with complex commands, the game is equipped with an interpreter which deciphers the input and maps it to in-game actions. For instance, in Figure~\ref{fig:zork1_egg}, a command ``climb the large tree" is issued, after which the player receives a response. In this example, the response explains that up in the tree is a collectible item - a jewel encrusted egg. The large, combinatorial action space is one of the main reasons Zork poses an interesting research problem. The actions are issued as free-text and thus the complexity of the problem grows exponentially with the size of the dictionary in use.


\textbf{Our setup:} In this work, we consider two tasks: the `Troll Quest' \citep{zahavy2018learn} and `Open Zork', i.e., solving the entire game. The `Troll Quest' is a sub-task within `Open Zork', in which the agent must enter the house, collect a lantern and sword, move a rug which reveals a trapdoor, open the trapdoor and enter the basement. Finally, in the basement, the agent encounters a troll which it must kill using the sword. An incorrect action at any stage may prevent the agent from reaching the goal, or even result in its death (termination).

In our setting, we consider a dictionary $D$ of $112$ unique words, extracted from a walk-through of actions which solve the game, a demonstrated sequence of actions (sentences) used to solve the game. We limit the maximal sentence length to $l=4$ words. Thus, the number of possible, unordered, word combinations are $d^l/l!$, i.e., the dictionary size to the power of the maximal sentence length, divided by the number of possible permutations. This results in approximately 10 million possible actions.


\textbf{Markov Decision Process (MDP):} Text-based games can be modeled as Markov Decision Processes.
An MDP $\mathcal{M}$ is defined by the tuple $(S, A, R, P)$ \citep{sutton1998reinforcement}. In the context of text-based games, the states $\state$ are paragraphs representing the current observation. $\aaa \in A$ are the available discrete actions, e.g., all combinations of words from the dictionary up to a maximal given sentence length $l$. $R : S \times A \times S \mapsto \mathbb{R}$ is the bounded reward function, for instance collecting items provides a positive reward. $P : S \times A \times S \mapsto [0, 1]$ is the transition matrix, where $P(\state'|\state,\aaa)$ is the probability of transitioning from $\state$ to $\state'$ given $\aaa$ was taken. 

\textbf{Action Space:} While the common approach may be to consider a discrete action space, such an approach may be infeasible to solve, as the complexity of solving the MDP is related to the effective action space size. Hence, in this work, we consider an alternative, continuous representation. As each action is a sentence composed of words, we represent each action using the sum of the embeddings of its tokens, or constitutive words, denoted by $\aae$ (Sum of Embeddings). A simple form of embedding is the BoW, it represents the word using a one-hot vector the size of the dictionary in which the dictionary index of the word is set to $1$. Aside from the BoW embedding, there exist additional forms of embedding vectors. For instance, Word2vec and GloVe, which encode the similarity between words (in terms of cosine distance). These embeddings are pre-trained using unsupervised learning techniques and similarly to how convolutional neural networks enable generalization across similar states, word embeddings enable generalization across similar sentences, i.e., actions. 


In this work, we utilize GloVe embeddings, pre-trained on the Wikipedia corpus. We chose GloVe over Word2vec, as there exist pre-trained embeddings in low dimensional space. The embedding space dimensionality is $m=50$, significantly smaller in dimension than the size $d$ of the dictionary $D$, $112$ in our experiments. Given the continuous representation of an action, namely the sum of embeddings of the sentence tokens $\aae \in \reals^m$, the goal is to recover the corresponding discrete action $\aaa$, that is the tokens composing the sentence. These may be represented as a BoW vector $\aab \in \integers^d$. Recovering the sentence from $\aab$ requires prior information on the language model.

Provided a set of words, the goal of a \emph{language model}, the last element in Figure~\ref{fig:algorithm}, a central piece in many important NLP tasks, is to output the most likely ordering which yields a grammatically correct sentence. In this paper, we use a rule based approach. Our rules are relatively simple. For example, given a verb and an object, the verb comes before the object - e.g., [`sword', `take'] $\mapsto$ `take sword'.

To conclude, we train a neural network $\text{E}_\theta (\state)$ to predict the sum of embeddings $\aae$. Using CS (Section~\ref{sec:compressed_sensing}), we recover the BoW vector $\text{R}(\aae) = \aab$, i.e., the set of words which compose the sentence. Finally, a language model M converts $\aab$ into a valid discrete-action, namely $\text{M}(\aab) = \aaa$. The combined approach is as follows:
    $\aaa = \text{M}(\text{R}(\text{E}(\state))) \enspace .$

\section{Compressed sensing}\label{sec:compressed_sensing}

This section provides some background on CS and sparse recovery, including practical recovery algorithms and theoretical recovery guarantees. In particular, we describe our variant of one popular reconstruction algorithm, OMP, that we refer to as Integer K-OMP (IK-OMP). The first modification allows exploitation of the integer prior on the sparse vector $\aab$ and is inspired by \citet{white2016generating} and \citet{sparrer2015soft}. The second mitigates the greedy nature of OMP using beam search \citep{lin2013kbest}. In \cref{sec: cs experiments}, we experimentally compare different sparse recovery methods and demonstrate the superiority of introducing the integer prior and the beam search strategy.


\subsection{Sparse Recovery}

CS is concerned with recovering a high-dimensional $p$-sparse signal $\xx \in \reals^d$ (the BoW vector $\aab$ in our setting) from a low dimensional measurement vector $\yy \in \reals^m$ (the sum of embeddings vector $\aae$). That is, given a dictionary $\DD \in \reals^{m \times d}$:
\begin{align} 
    \min ||\xx||_0  \enspace \text{subject to} \enspace \DD\xx = \yy.
    \label{eq:cs_l0}
\end{align}

To ensure uniqueness of the solution of \eqref{eq:cs_l0}, the sensing matrix, or dictionary, $\DD$
must fulfill certain properties. These are key to provide practical recovery guarantees as well. Well known such properties are the spark, or Kruskal rank \citep{donoho2003optimally}, and the Restricted Isometry Property (RIP) \citep{candes2005decoding}. Unfortunately, these are typically as hard to compute as solving the original problem \eqref{eq:cs_l0}. While the mutual-coherence (see Definition~\ref{def:coherence}) provides looser bounds, it is easily computable. Thus, we focus on mutual-coherence based results and note that Spark and RIP based guarantees may be found in \cite{elad2010book}.


\begin{definition}[\cite{elad2010book} Definition 2.3]\label{def:coherence}
    The mutual coherence of a given matrix $\DD$ is the largest absolute normalized inner product between different columns from $\DD$. Denoting the $k$-th column in $\DD$ by $\dd_k$, it is given by
        $\mu (\DD) = \max_{1 \leq i, j \leq m, \enspace i \neq j} \frac{| \dd_j^T \dd_i |}{||\dd_i||_2 ||\dd_j||_2}.$
\end{definition}

The mutual-coherence characterizes the dependence between columns of the matrix $\DD$. For a unitary matrix, columns are pairwise orthogonal, and as a result, the mutual-coherence is zero. For general matrices with more columns than rows ($m < d$), as in our case, $\mu$ is necessarily strictly positive, and we desire the smallest possible value so as to get as close as possible to the behavior exhibited by unitary matrices \citep{elad2010book}. This is illustrated in the following uniqueness theorem.

\begin{theorem}[\cite{elad2010book} Theorem 2.5]
\label{th:cs_unique}
If a system of linear equations $\DD\xx=\yy$ has a solution $\xx$ obeying 
        $p < \frac{1}{2} \left( 1 + \frac{1}{\mu(\DD)} \right),$
where $p=||\xx||_0$, this solution is the sparsest possible.
\end{theorem}
We now turn to discuss practical methods to solve \eqref{eq:cs_l0}.

\subsection{Recovery Algorithms}

The sparse recovery problem \eqref{eq:cs_l0} is non-convex due to the $\ell_0$-norm. Although it may be solved via combinatorial search, the complexity is exponential in the dictionary dimension $d$, and it has been proven that \eqref{eq:cs_l0} is, in general, NP-Hard \citep{elad2010book}.\\
One approach to solve \eqref{eq:cs_l0}, \textit{basis pursuit}, relaxes the $\ell_0$-minimization to its $\ell_1$-norm convex surrogate, 
\begin{align} \label{eq:cs_l1}
    \min ||\xx||_1\enspace \text{s.t.} \enspace \DD\xx = \yy.
\end{align}
In the presence of noise, the condition $\DD\xx=\yy$ is replaced by $||\DD\xx-\yy||_2 \leq \epsilon$. The Lagrangian relaxation of this quadratic program is written, for some $\lambda>0,$ as
    $\min ||\xx||_1 + \lambda ||\yy-\DD\xx||_2,$
and is known as basis pursuit denoising (BPDN).

The above noiseless and noisy problems can be respectively cast as linear programming and second order cone programming problems \citep{chen2001atomic}. They thus may be solved using techniques such as interior-point methods \citep{ben2001lectures, boyd2004convex}. Large scale problems involving dense sensing matrices often precludes the use of such methods. This motivated the search for simpler gradient-based algorithms for solving \eqref{eq:cs_l1}, such as fast iterative shrinkage-thresholding algorithm (FISTA) \citep{beck2009fast}.


Alternatively, one may use greedy methods, broadly divided into \textit{matching pursuit} based algorithms, such as OMP \citep{blumensath2008gradient}, and \textit{thresholding} based methods, including iterative hard thresholding \citep{blumensath2009iterative}. The popular OMP algorithm, proceeds by iteratively finding the dictionary column with the highest correlation to the signal residual, computed by subtracting the contribution of a partial estimate of $\xx$ from $\yy$. The coefficients over the selected support set are then chosen so as to minimize the residual error. A typical halting criterion compares the residual to a predefined threshold.

\subsection{Recovery Guarantees}
Performance guarantees for both $\ell_1$-relaxation and greedy methods have been provided in the CS literature. In noiseless settings, under the conditions of Theorem~\ref{th:cs_unique}, the unique solution of \eqref{eq:cs_l0} is also the unique solution of \eqref{eq:cs_l1} \citep[Theorem 4.5]{elad2010book}. Under the same conditions, OMP with halting criterion threshold $\epsilon = 0$ is guaranteed to find the exact solution of \eqref{eq:cs_l0} \citep[Theorem 4.3]{elad2010book}. More practical results are given for the case where the measurements are contaminated by noise \citep{donoho2006stable,elad2010book}.

\subsection{Integer K-OMP (IK-OMP)}

\begin{algorithm}[H]
  \caption{IK-OMP}\label{algo:beam_omp}
  \begin{algorithmic}
    \State \textbf{Input:} Measurement vector $\yy \in \reals^m$, dictionary $\DD \in \reals^{m \times d}$, maximal number of characters $L$ and beam width $K$
    \State Initial solutions $\XX^0 = [\mathbf{0}_d, \dots, \mathbf{0}_d]$
    \For{$l = 1, L$}
        \For{$i \in [1, \dots, k]$}
            \State \textbf{Extend:} Append $\XX_i^{l-1}\! +\! \textbf{1}_j,\! \forall j\!\! \in\! [1, ..., d]$ to $\XX^{l-1}$
        \EndFor
        \State \textbf{Trim:} $\XX^l= \text{K-}\argmin_{\XX_i \in \XX^{l-1}} ||\yy - \DD\XX_i||_2^2$
    \EndFor
    \State \textbf{return} $\XX^L$
  \end{algorithmic}
\end{algorithm}

\textbf{An Integer Prior:} While CS is typically concerned with the reconstruction of a sparse real-valued signal, in our BoW linear representation, the signal fulfills a secondary structure constraint besides sparsity. Its nonzero entries stem from a finite, or discrete, alphabet. Such prior information on the original signal appears in many communication scenarios \citep{candes2005error, axell2012spectrum}, where the transmitted data originates from a finite set.

\textbf{Beam Search OMP:} As OMP iteratively adds atoms to the recovered support, the choice of a new element in an iteration is blind to its effect on future iterations. Therefore, any mistakes, particularly in early iterations, may lead to large recovery errors. To mitigate this phenomenon, several methods have been proposed to amend the OMP algorithm. 

To decrease the greediness of the greedy addition algorithm (which acts similarly to OMP), \citet{white2016generating} use a substitution based method, also referred as swapping \citep{andrle2006swapping} in the CS literature. Unfortunately, the computational complexity of this substitution strategy makes it impractical. \citet{elad2009plurality} combine several recovered sparse representations, to improving denoising, by randomizing the OMP algorithm. However, in our case, the sum of embeddings $\aae$ represents a true sparse BoW vector $\aab$, so that combining several recovered vectors should not lead to the correct solution.

\textbf{IK-OMP:} We combine the integer-prior with the beam search strategy, and propose the IK-OMP (Algorithm~\ref{algo:beam_omp}). In the algorithm description, $\mathbf{1}_j$ is the vector with a single nonzero element at index $j$ and $\text{K-}\argmin$ denotes the $K$ elements with smallest value for the following expression. In this work, the selected BoW is the candidate which minimizes the reconstruction score.

\begin{figure*}[t]
\centering
Troll Quest \hspace{7cm} Open Zork \\
\begin{subfigure}{.24\textwidth}
    \centering
	\includegraphics[width=\linewidth]{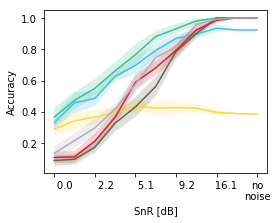}
\end{subfigure}%
\begin{subfigure}{.24\textwidth}
    \centering
	\includegraphics[width=\linewidth]{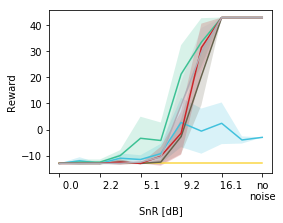}
\end{subfigure}%
\begin{subfigure}{.24\textwidth}
    \centering
	\includegraphics[width=\linewidth]{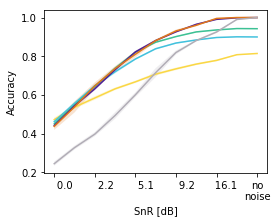}
\end{subfigure}%
\begin{subfigure}{.24\textwidth}
    \centering
	\includegraphics[width=\linewidth]{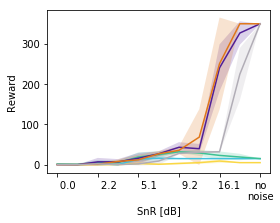}
\end{subfigure}%
\\
\hspace{0.3cm}
\begin{subfigure}{.43\textwidth}
    \centering
	\includegraphics[width=\linewidth]{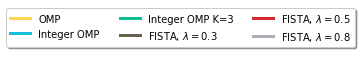}
\end{subfigure}%
$\enspace\enspace\enspace\enspace$
\begin{subfigure}{.45\textwidth}
    \centering
	\includegraphics[width=\linewidth]{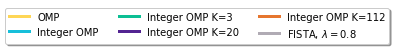}
\end{subfigure}%
\caption{\textbf{Compressed Sensing}: Comparison of the accuracy, and accumulated reward, of the various reconstruction algorithms on the `Troll Quest' and in `Open Zork'.
The SnR denotes the ratio between the norm of the original signal $\aae$ and that of the added noise.}
\label{fig:full compressed sensing}
\vspace{-0.3cm}
\end{figure*}

\section{Imitation Learning}
\label{sec:imitationlearning}
In this section, we present our \alg\ algorithm and provide in-depth details regarding the design and implementation of each of its underlying components. We also detail the experiments of executing \alg\ on the entire game of Zork.

\textbf{\algfull:} Our \alg\ architecture is composed of two major components - Encoder $\text{E}_\theta (\state)$ and Retrieval Mechanism (as seen in Figure~\ref{fig:algorithm}). Each component has a distinct role and combining them together enables for a computationally efficient approach. 

\textbf{The Encoder (E)} is a neural network trained to output the optimal action representation at each state. As we consider the task of imitation learning, this is performed by minimizing the $\ell_2$ loss between the Encoder's output $E_{\theta}(\state)$ and the embedding of the action provided by the expert $\aae$.

\begin{figure*}[t]
    \hspace{-0.5cm}
    \parbox[t]{0.27\textwidth}{\null
        \centering
        \vspace{0.4cm}
        \includegraphics[width=\linewidth]{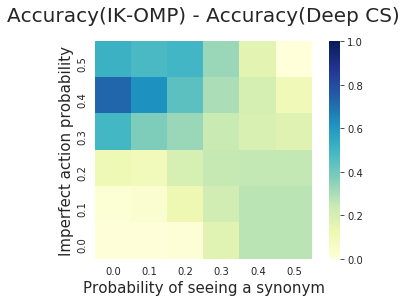}
        \caption{Difference in reconstruction accuracy, between \textbf{Sparse-IL and DeepCS-2}. Higher value represents a higher reconstruction accuracy for Sparse-IL. DeepCS-2 fails when presented with several variants of the correct actions (synonyms).}
        \label{fig:deepcs vs ik-omp accuracy comparison}
    }
    \hspace*{0.49cm}
    \parbox[t]{0.73\textwidth}{\null
        \centering
        Open Zork\\
        \begin{subfigure}{.24\textwidth}
            \centering
        
        	\includegraphics[width=\linewidth]{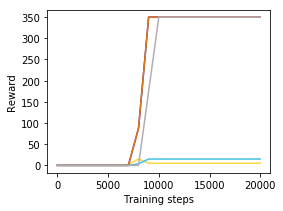}
        \end{subfigure}%
        \begin{subfigure}{.24\textwidth}
            \centering
        	\includegraphics[width=\linewidth]{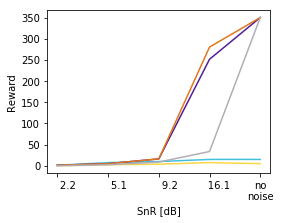}
        \end{subfigure}%
        \begin{subfigure}{.24\textwidth}
            \centering
        
        	\includegraphics[width=\linewidth]{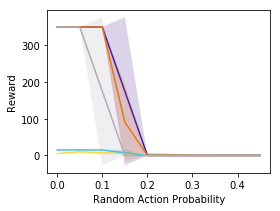}
        \end{subfigure}%
        \\
        \begin{subfigure}{.6\textwidth}
            \centering
        	\includegraphics[width=\linewidth]{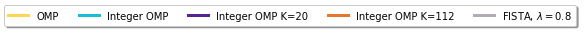}
        \end{subfigure}%
        
        \caption{\textbf{Sparse Imitation Learning:} Comparison of the accuracy of each reconstruction algorithm on an agent trained using imitation learning to solve the entire game. In the graph on the left, IK-OMP with K=20 and K=112 result in identical performance.}
        \label{fig:openzork imitation learning}
    }
\end{figure*}

In all of the learning experiments, the architecture we use is a convolutional neural network (CNN) that is suited to NLP tasks \citep{kim2014convolutional}. Due to the structure of the game, there exist long term-dependencies. Frame-stacking, a common approach in games \citep{mnih2015human}, tackles this issue by providing the network with the N previous states. For the ``Open Zork'' task, we stack the previous 12 states, whereas for the ``Troll Quest'' we only provide it with the current frame.

\textbf{Retrieval Mechanism (R):} The output of the Encoder, $\text{E}_{\theta} (\state)$, is fed into a CS algorithm, such as IK-OMP. IK-OMP produces K candidate actions, ${\aab}_1, ..., {\aab}_K$. These actions are fed into a fitness function which ranks them, based on the reconstruction score $||\text{E}_{\theta} (\state) - \DD{\aab}_i||_2^2 , \enspace \forall i = 1, ..., k$ (see Section~\ref{sec:compressed_sensing}), and returns the optimal candidate. Other CS approaches, e.g., OMP and FISTA, return a single candidate action.

\section{Experiments}\label{sec: experiments}

In this section, we present our experimental results. We begin by analyzing our proposed CS method, namely IK-OMP, in \cref{sec: cs experiments}, and its ability to reconstruct the action when provided the sum of word embeddings $\aae$. After evaluating our proposed method in a clean and analyzable scenario, we evaluate the entire system `\algfull'\ on the full game of Zork (\cref{sec: il experiments}).

\subsection{Compressed Sensing}\label{sec: cs experiments}
In this section, we focus on comparing several CS approaches. To do so, we follow the set of commands, extracted from a walk-through of the game, required to solve Zork1, both in the `Troll Quest' and `Open Zork' domains. In each state $\state$, we take the ground-truth action $\aaa (\state)$, calculate the sum of word embeddings $\aae (\state)$, add noise and test the ability of various CS methods to reconstruct $\aaa (\state)$. We compare the \emph{run-time} (Table~\ref{table:sparse comparison}), and the \emph{reconstruction accuracy} (number of actions reconstructed correctly) and \emph{reward gained} in the presence of noise (Figure~\ref{fig:full compressed sensing}). Specifically, the measured action is $\mathbf{a}_\text{mes} (\state) = \aae (\state) + \epsilon$, where $\epsilon \sim N(0,1)$ is normalized based on the signal to noise ratio (SnR).

\begin{table}[H]
    \caption{Runtime comparison.}\label{table:sparse comparison}
    \centering
    \begin{tabular}{|l|l|}
    	\hline
        \\[-1em]
        \textbf{Algorithm} & \textbf{Runtime} \\
        \hline
        \\[-1em]
        OMP & 0.008 \\
        \hline
        \\[-1em]
        IOMP, K=1 & 0.008 \\
        \hline
        \\[-1em]
        IK-OMP, K=3 & 0.021 \\
        \hline
        \\[-1em]
        IK-OMP, K=20 & 0.166 \\
        \hline
        \\[-1em]
        IK-OMP, K=112 & 1.116 \\
        \hline
        \\[-1em]
        FISTA &0.881 \\
        \hline
        \\[-1em]
        DeepCS &0.347 \\
        \hline
    \end{tabular}
\end{table}

We compare 4 CS methods: the FISTA implementation of BP, OMP, IK-OMP (Algorithm~\ref{algo:beam_omp}) and a Deep Learning variant we deem DeepCS described below. The dictionary is composed of $d=112$ possible words which can be used in the game. The dimension of the embedding is $m=50$ (standard GloVe embedding available online) and the sentence length is limited to at most $4$ words. This yields a total number of $\approx$ 10 million actions, from which the agent must choose one at each step. It is important to note that while \emph{accuracy} and \emph{reward} might seem similar, an inaccurate reconstruction at an early stage results in an immediate failure, even when the accuracy over the entire trajectory seems high.

Clearly, as seen from Figure~\ref{fig:full compressed sensing}, OMP fails to reconstruct the true BoW vectors $\aab$, even in the noiseless scenario. Indeed, the mutual-coherence (Definition~\ref{def:coherence}) is $\mu=0.97$ and from Theorem~\ref{th:cs_unique}, there is no guarantee that OMP can reconstruct a sparse vector for any sparsity $p>0$. However, our suggested approach, IK-OMP, is capable of correctly reconstructing the original action $\aab$, even in the presence of relatively large noise. This gives evidence that the integer prior, in particular, and the beam search strategy significantly improve the sparse recovery performance.

\textbf{Deep Compressed Sensing:} Besides traditional CS methods, it is natural to test the ability of deep learning methods to perform such a task. Here, we train a neural network to predict the BoW vector $\aab$ which composes the continuous embedding vector. Our network is a multi layer perceptron (MLP), composed of two hidden layers, 100 neurons each. We use a sigmoid activation function to bound the outputs to $[0,1]$ and train the network using a binary cross entropy loss. In these experiments, we denote by $T$ the threshold above which an output is selected, e.g., when $T=0.9$ all words which receive a weight of above $0.9$ are selected.

\begin{figure}[H]
    \centering
    Open Zork \\
    \begin{subfigure}{.24\textwidth}
        \centering
    	\includegraphics[width=\linewidth]{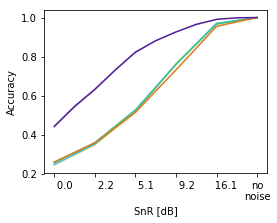}
    \end{subfigure}%
    \begin{subfigure}{.24\textwidth}
        \centering
    	\includegraphics[width=\linewidth]{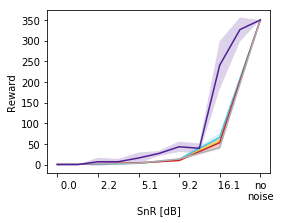}
    \end{subfigure}%
    \\
    \begin{subfigure}{.45\textwidth}
        \centering
    	\includegraphics[width=\linewidth]{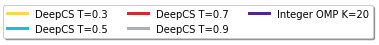}
    \end{subfigure}%
    \caption{\textbf{Compressed Sensing - DeepCS:} Comparison of the accuracy, and accumulated reward, of the DeepCS baselines, compared to the IK-OMP approach.}
    \label{fig:compressed sensing deep cs openzork}
\end{figure}

Our results (Figure~\ref{fig:compressed sensing deep cs openzork}) show that the DeepCS approach works when no noise is present, however, once noise is added to the setup, it is clear that DeepCS performs poorly  compared to classic CS methods such as IK-OMP. We observed similar results in the Troll domain. Besides, as DeepCS requires training a new model for each domain, it is data-specific and does not transfer easily, which is not the case with traditional CS methods. 

\subsection{Imitation Learning}\label{sec: il experiments}


In an imitation learning setup, we are given a data set of state-action pairs $(\state, \aaa)$, provided by an expert; the goal is to learn a policy that achieves the best performance possible. We achieve this by training the embedding network $\text{E}_\theta (\state)$ to imitate the demonstrated actions in the embedding space, namely $\aae$, at each state $\state$, using the MSE between the predicted actions and those demonstrated. We consider three setups: (1) Perfect demonstrations, where we test errors due to architecture capacity and function approximation, (2) Gaussian noise, $\mathbf{a}_\text{mes} (\state) = \aae (\state) + \epsilon$ (See Section~\ref{sec: cs experiments}), and (3) discrete-action noise, in which a random incorrect action is demonstrated with probability (w.p.) $p$. 
This experiment can be seen as learning from demonstrations provided by an ensemble of sub-optimal experts.

Our results (Figure~\ref{fig:openzork imitation learning}) show that by combining CS with imitation learning techniques, we are capable of solving the entire game of Zork1, even in the presence of discrete-action noise. In all our experiments, IK-OMP outperforms the various baselines, including the end-to-end approach DeepCS-2 which is trained to predict the BoW embedding $\aab$ directly from the state $\state$.

\textbf{Training:} Analyzing the training graph presents an interesting picture. It shows that during the training process, the output of the Encoder can be seen as a noisy estimation of $\aae$. As training progresses, the effective SnR of the noise decreases which is seen by the increase in the reconstruction performance.

\textbf{Generalization:} In Figure~\ref{fig:deepcs vs ik-omp accuracy comparison}, we present the generalization capabilities which our method \alg\ enjoys, due to the use of pre-trained unsupervised word embeddings. The heatmap shows two forms of noise. The first, as before, is the probability of receiving a bad demonstration, an incorrect action. The second, synonym probability, is the probability of being presented with a correct action, yet composed of different words, e.g., drop, throw and discard result in an identical action in the environment and have a similar meaning. These results clearly show that Sparse-IL outperforms DeepCS-2 in nearly all scenarios, highlighting the generalization improvement inherent in the embeddings.


\textbf{The benefit of meaningful embeddings:} In our approach, the Encoder $\text{E}_\theta$ is trained to predict the sum-of-embeddings $\aae$. However, it can also be trained to directly predict the BoW vector $\aab$. While this approach may work, it lacks the generalization ability which is apparent in embeddings such as GloVe, in which similar words receive similar embedding vectors.

Consider a scenario in which there are 4 optimal actions (e.g., `go north', `walk north', 'run north' and `move north') and 1 sub-optimal action (e.g., `climb tree'). With probability $0.15$ we are presented with one of the optimal actions and with probability $0.4$ the sub-optimal action. In this example, the expected BoW representation would include `north' w.p. $0.6$, `climb' and `tree' w.p. $0.4$, and the rest w.p. $0.15$. On the other hand, since `go', `walk', `run' and `move' have similar meanings and in turn similar embeddings, the expected $\aae$ is much closer to the optimal actions than to the sub-optimal one and thus an imitation agent is less likely to make a mistake. 

\section{Conclusion}\label{sec:conclusion}
We have presented a computationally efficient algorithm called \algfull\ (\alg) that combines CS with imitation learning to solve text-based games with combinatorial action spaces. We proposed a CS algorithm variant of OMP which we have called Integer K-OMP (IK-OMP) and demonstrated that it can deconstruct a sum of word embeddings into the individual BoW that make up the embedding, even in the presence of significant noise. In addition, IK-OMP is significantly more computationally efficient than the baseline CS techniques. When combining IK-OMP with imitation learning, our agent is able to solve Troll quest as well as the entire game of Zork1 for the first time. Zork1 contains a combinatorial action space of 10 million actions. Future work includes replacing the fitness function with a critic in order to further improve the learned policy as well as testing the capabilities of the critic agent in cross-domain tasks.


\bibliographystyle{plainnat}
\bibliography{main.bib}

\end{document}